\documentclass[letterpaper]{article} 
\usepackage[draft]{aaai2026}  
\usepackage{times}  
\usepackage{helvet}  
\usepackage{courier}  
\usepackage[hyphens]{url}  
\usepackage{graphicx} 
\usepackage{natbib}  
\usepackage{caption} 
\usepackage{algorithm}
\usepackage{algorithmic}

\usepackage{newfloat}
\usepackage{listings}
\DeclareCaptionStyle{ruled}{labelfont=normalfont,labelsep=colon,strut=off} 
\floatstyle{ruled}
\newfloat{listing}{tb}{lst}{}
\floatname{listing}{Listing}
\usepackage{amsfonts}
\usepackage{amsmath}
\usepackage{amssymb} 
\usepackage{booktabs}
\usepackage{multirow} 
\usepackage{times}
\usepackage{helvet}
\usepackage{courier}
\usepackage[table]{xcolor}
\usepackage{fancyvrb}

\title{CAD-Judge: Toward Efficient Morphological Grading and Verification for Text-to-CAD Generation}
\author{
Zheyuan Zhou\textsuperscript{\rm 1}\equalcontrib,
Jiayi Han\textsuperscript{\rm 2}\equalcontrib,
Liang Du\textsuperscript{\rm 3},
Naiyu Fang\textsuperscript{\rm 4},
Lemiao Qiu\textsuperscript{\rm 1},
Shuyou Zhang\textsuperscript{\rm 1}
}
\affiliations{
    \textsuperscript{\rm 1} Zhejiang University \\
    \textsuperscript{\rm 2} Inspur Genersoft Co. Ltd., Inspur Group Co. Ltd. \\
    \textsuperscript{\rm 3} Tencent Inc.\\
    \textsuperscript{\rm 4} Nanyang Technological University\\[8pt]
    \textcolor{cyan}{\url{https://zhouzheyuan.github.io/cad-judge}}
}

\usepackage{bibentry}

\begin{document}

\maketitle

\begin{abstract}
Computer-Aided Design (CAD) models are widely used across industrial design, simulation, and manufacturing processes. 
Text-to-CAD systems aim to generate editable, general-purpose CAD models from textual descriptions, significantly reducing the complexity and entry barrier associated with traditional CAD workflows. 
However, rendering CAD models can be slow, and deploying VLMs to review CAD models can be expensive and may introduce reward hacking that degrades the systems. 
To address these challenges, we propose CAD-Judge, a novel, verifiable reward system for efficient and effective CAD preference grading and grammatical validation. 
We adopt the Compiler-as-a-Judge Module (CJM) as a fast, direct reward signal, optimizing model alignment by maximizing generative utility through prospect theory. 
To further improve the robustness of Text-to-CAD in the testing phase, we introduce a simple yet effective agentic CAD generation approach and adopt the Compiler-as-a-Review Module (CRM), which efficiently verifies the generated CAD models, enabling the system to refine them accordingly. 
Extensive experiments on challenging CAD datasets demonstrate that our method achieves state-of-the-art performance while maintaining superior efficiency.
\end{abstract}
\section{Introduction}
Computer-Aided Design (CAD) has become an essential tool in modern industrial design and manufacturing, enabling the precise modeling of complex geometries~\cite{robertson1993cad, cherng1998feature, yamamoto2005interaction, deng2024sets, khan2024cad}. Traditional workflows involve manually creating 2D sketches using basic primitives such as lines, arcs, and circles, which are then extruded into 3D solid models. While this process is powerful, it demands significant expertise and iterative refinement, often limiting accessibility for non-expert users.

To democratize CAD modeling, recent efforts have focused on text-to-CAD systems—approaches that aim to automate the generation of parametric command sequences directly from natural language descriptions~\cite{text2cad, CADTranslator, cadfusion}. Such systems allow both professionals and hobbyists to rapidly prototype editable CAD models through intuitive text instructions, significantly reducing the time and effort required for manual construction. Current approaches typically rely on end-to-end Transformer-based architectures trained on paired datasets of textual commands and corresponding CAD sequences~\cite{CADTranslator, text2cad}. These models are often trained from scratch, resulting in slow convergence and suboptimal performance. Moreover, limited model capacity restricts their ability to interpret nuanced or complex design instructions. Recent works attempt to address this by integrating pre-trained large language models (LLMs), leveraging their strong language understanding capabilities and general knowledge of CAD principles~\cite{cadfusion}.

\begin{figure*}[t]
\centering
\includegraphics[width=0.85\linewidth]{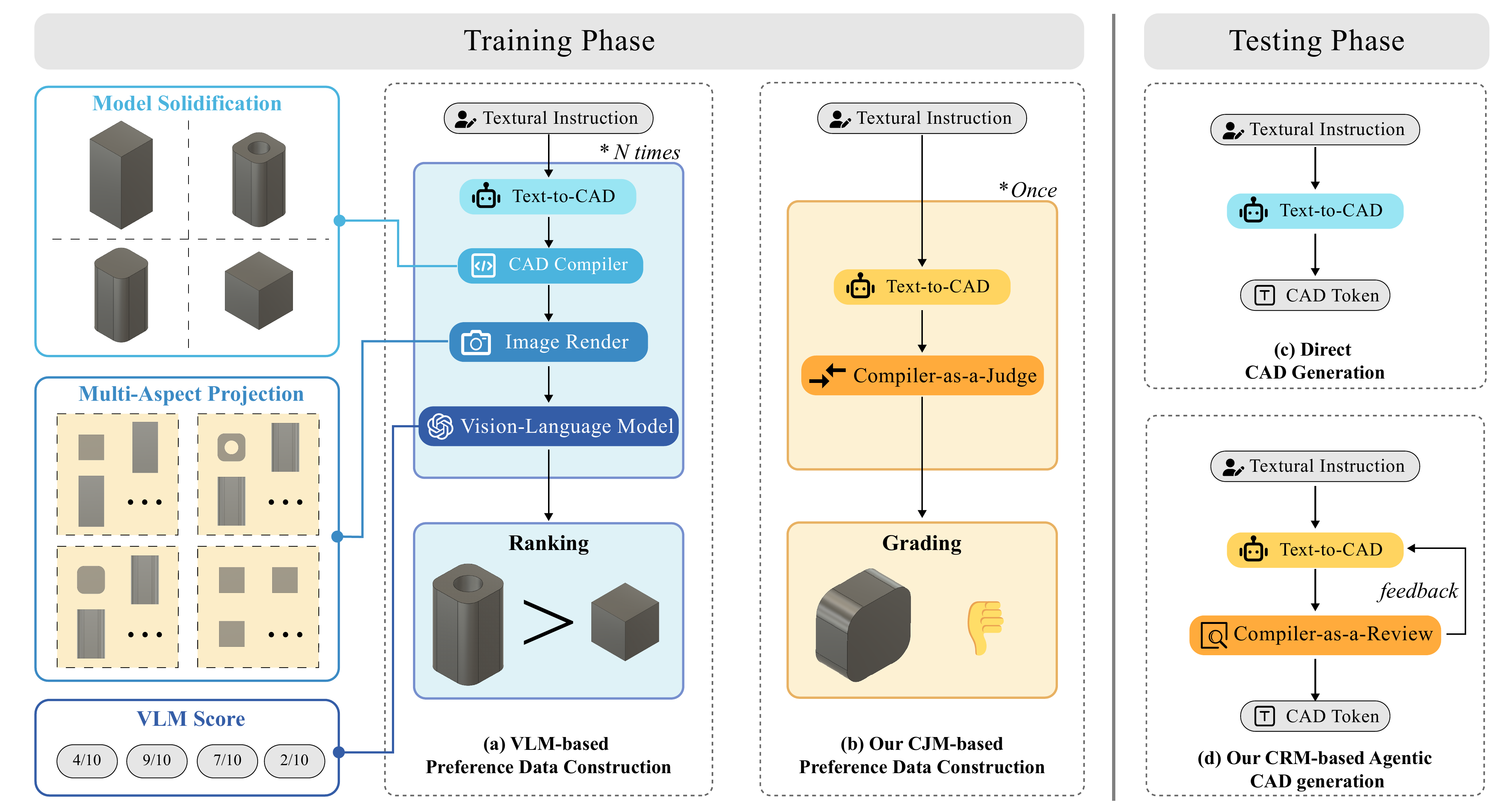}
\caption{Comparison between existing preference data construction pipelines and ours.
(a) Existing Vision-Language Model (VLM)-based methods rely on costly multi-sample rendering and projection for score grading.
(b) Our method leverages per-sample preferences derived from a native quantifiable CAD compiler, enabling efficient and scalable data collection.
The ``\(>\)'' symbol denotes pairwise preferences, where one item is selected as superior from a ranked list, while the thumbs-down icon indicates independent binary negative preferences without explicit pairing.
(c) Existing direct CAD sequence generation methods output the final CAD sequence directly through LLMs.
(d) Our approach incorporates feedback on erroneous samples, enabling the model to correct obvious mistakes and iteratively generate the final CAD sequence.
}
\label{fig:teaser}
\end{figure*}
However, aligning these models with human preferences remains a major hurdle. 
As shown in Figure~\ref{fig:teaser}, most existing methods depend on pairwise preference data—comparisons between preferred and rejected CAD outputs generated from the same prompt~\cite{cadfusion}.
Constructing such data is costly, requiring rendering and ranking multiple generations per prompt via VLMs. 
Additionally, LLM-generated CAD sequences often fail to conform to strict syntactic rules, leading to high invalidation rates during compilation.

To address these limitations, we introduce \textbf{CAD-Judge}, a novel verifiable reward system for efficient and effective CAD preference grading and grammatical validation. Unlike conventional methods that rely on Vision-Language Models (VLMs) and their dependence on rendered images, our approach leverages per-sample binary feedback obtained via a Compiler-as-a-Judge Module (CJM). Specifically, our CJM employs the Chamfer Distance (CD) between generated and ground-truth CAD models as a fast, interpretable reward signal—obviating the need for computationally costly rendering and ranking procedures. Furthermore, to improve the robustness of LLM-generated sequences, we introduce the Compiler-as-a-Review Module (CRM), which efficiently verifies the generated CAD models, allowing the LLMs to modify them. By monitoring compilation outcomes and performing intelligent adjustments based on error diagnostics, this scaling strategy significantly boosts the likelihood of generating valid, executable CAD programs without incurring substantial computational overhead.

Our key contributions are summarized as follows:
\begin{itemize}
\item An efficient Compiler-as-a-Judge Module (CJM) for constructing per-sample binary preference data using CJ-derived rewards.
\item A paradigm grounded in prospect theory that directly optimizes the alignment of generated CAD command sequences by maximizing generative utility.
\item A Compiler-as-a-Review Module (CRM) that efficiently reviews the LLM-generated CAD sequences during test time for robust CAD generation.
\item Extensive experimental evaluations on various datasets demonstrating that CAD-Judge achieves state-of-the-art performance on text-to-CAD tasks while maintaining high computational efficiency.
\end{itemize}

\section{Related Workers}

\textbf{CAD Generation.}
CAD, depending on its representation methods, can be categorized into constructive solid geometry (CSG)~\cite{du2018inversecsg, kania2020ucsg, yu2022capri, yu2023d}, boundary representation (B-rep)~\cite{jayaraman2022solidgen,wang2022neural,xu2024brepgen}, and sketch-and-extrude modeling (SEM)~\cite{willis2021engineering, deepcad}. 
Our research primarily focuses on SEM, which records the drawing history of a CAD model, starting from curves, progressing to 2D sketches, and ultimately culminating in the extrusion to form a 3D model.
DeepCAD~\cite{deepcad} was the pioneer in proposing a sketch-extrusion construction sequence representation for CAD models. In a preliminary experiment, it predicted the CAD history from latent vectors or point clouds.
CAD Translator~\cite{CADTranslator} utilize an encoder-decoder architecture. 
By aligning the encoded CAD Embedding with the Text Embedding, the model acquires the ability to generate CAD sequences from text. 
After the fusion process, the embeddings are fed into the decoder to recover 3D parametric CAD sequences.
Text2CAD~\cite{text2cad} trains an end-to-end transformer-based autoregressive network to generate parametric CAD models from input texts. 
~\cite{yavartanoo2024text2cad} employs images as an intermediate medium and realizes the generation in multiple steps. Specifically, the text is transformed into an isometric image that represents the described features. 
Subsequently, this image is mapped into orthographic technical drawings, which serve as the basis for generating the 3D CAD model.
CAD-GPT~\cite{cadgpt} introduces a Multimodal Large Language Model (MLLM) that can precisely synthesize CAD modeling sequences from either a single image or a textual description. The CAD-MLLM has the capability to generate parametric CAD models based on multimodal inputs, such as text, images, point clouds, or a combination of these modalities.
CAD-Coder~\cite{cad-coder} fine-tunes LLMs to generate CadQuery scripts for building CAD models by incorporating GRPO and CoT techniques.
CADmium~\cite{cadmium} introduces a novel annotation pipeline that leverages GPT-4.1 to process up to 10 multi-view images of 3D objects along with their construction sequences.
CADFusion~\cite{cadfusion} leverages 2D renderings generated by generative models and utilizes the image knowledge from a Vision-Language Model (VLM) as a reward signal to construct a DPO dataset, improving the performance of the model after SFT through an iterative training procedure. 
In contrast, our method does not rely on DPO, and therefore does not require pairwise preference annotations. 
By only requiring binary preferences, our approach enables the rapid and efficient construction of high-quality alignment data without the need for LVMs.

\textbf{Large Language Models (LLMs).}
LLMs have achieved remarkable success in recent years, primarily through the pretraining process on extensive general-purpose datasets.
To adapt LLMs to specific tasks, supervised fine-tuning (SFT) has emerged as a powerful approach. 
By training LLMs on task-specific labeled data, SFT can significantly enhance the models' performance on targeted tasks \cite{chen2024huatuogpto1medicalcomplexreasoning,yue2024lawllm,han-etal-2025-slim,liu2025finr1largelanguagemodel}.
Reinforcement Learning (RL) is another crucial technique to align LLM outputs with human preferences. 
PPO \cite{ppo} aims to iteratively update the model's policy to maximize cumulative rewards based on human feedback or predefined evaluation metrics. 
DPO \cite{dpo} simplifies the RL process by directly optimizing the model's output based on pairwise human preferences, reducing computational complexity while maintaining effectiveness. 
\begin{figure*}[t]
\centering
\includegraphics[width=0.9\linewidth]{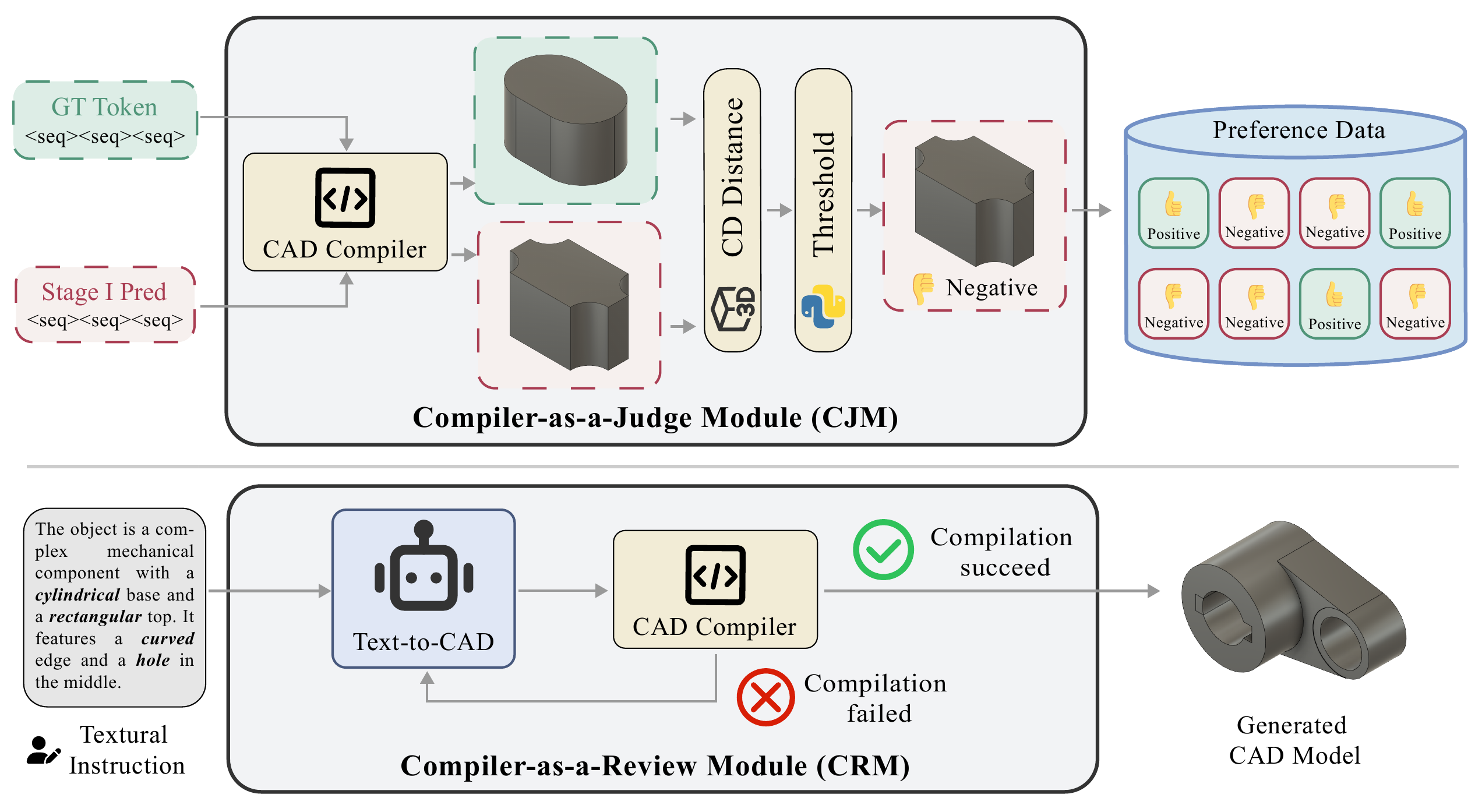}
\caption{Overview of our CAD-Judge framework.
Our \textbf{Compiler-as-a-Judge Module (CJM)} functions as a rule-based reward signal for grading after the Stage I SFT training.
Predictions exceeding the Chamfer Distance (CD) threshold are treated as negative samples, while those with CD below the threshold undergo are selected with as positive samples.
During inference, our \textbf{Compiler-as-a-Review Module (CRM)} validates whether the generated CAD sequence can be successfully compiled. 
If compilation fails, the model resamples until a valid sequence is produced, thereby significantly reducing the invalid generation ratio.}
\label{fig:framework}
\end{figure*}
Resently, KTO \cite{kto} directly maximizes the utility of generations instead of maximizing the log-likelihood of preferences, which only requires a binary signal of whether an output is desirable or undesirable for an input.
Many recent works propose using the rule-based approach to identify the samples to be chosen or rejected to avoid reward hacking. 
For example, DeepSeek-R1 proposes GPRO \cite{deepseekai2025deepseekr1incentivizingreasoningcapability}, which leverages rule-based reward, could significantly enhance the reasoning capacity of LLMs. 
Visual-RFT \cite{liu2025visualrftvisualreinforcementfinetuning} also proposes to utilize rule-based metrics (e.g., F1-score, IoU) to sample preferred reasoning trajectories to enhance multi-model perception tasks. 
\section{Methodology}
\subsection{Overview}
Let a textual description be denoted as $x$, and a CAD parametric sequence as $y$. 
The text-to-CAD task can be formulated as first encoding the description text $x$ into a CAD parametric representation $y$ using a function $f(\cdot)$.
The generated CAD parametric sequence $y$ can then be rendered into a 3D solid object via a CAD compiler $c(\cdot)$, such as \texttt{pythonOCC}~\cite{pythonocc}.
The overall objective is for the rendered 3D model $m = c(f(x))$, to closely match the user's intended 3D object.

Our CAD-Judge introduces a two-stage training paradigm that integrates supervision with performance feedback for text-to-CAD generation. 
In the first stage, a pre-trained LLM is fine-tuned using text-CAD sequence pairs as supervised signals, enabling the model to learn the structural patterns and compositional logic of CAD sequences while adapting its general language capabilities to the domain-specific syntax of CAD instructions. 
The second stage incorporates compiler-mediated feedback, where the LLM is incentivized to generate executable and geometrically accurate sequences by rewarding successful compilations and penalizing errors or deviations from ground-truth models. 
This dual-objective approach bridges symbolic reasoning with functional execution constraints, enhancing both syntactic correctness and geometric fidelity. 

\subsection{Agentic CAD Generation}
Despite their strong language understanding capabilities, LLMs often produce outputs that exhibit syntactic or structural inconsistencies when applied to domain-specific tasks such as text-to-CAD generation. 
Prior work has shown that LLM-generated CAD sequences suffer from a high invalidation rate due to deviations from the expected parametric command structure \cite{CADTranslator, cadfusion}. 
These irregularities prevent successful compilation into executable CAD models, limiting the practical usability of such systems.
To mitigate this issue, we propose a simple yet effective agentic CAD generation approach. There are two nodes in this agent: the generation node and the review node. The generation node is implemented with a well-trained LLM (specifically, the model with our two-stage fine-tuning with verifiable CAD reward), and the review node is implemented with CRM. During inference, the generation node first generates the CAD sequence according to the input prompt. The review node verifies the validation of the generated sequence by CRM, and expands the review information to the input prompt for further refinement. We iterate this generate-review phase until the sequence is valid or exceeds the iteration limit. 

\subsection{Compiler-as-a-Judge Module (CJM) for Verifiable CAD Reward}

The core objective of Text-to-CAD task is to ensure that 3D CAD models accurately align with the design intent described in text while strictly adhering to the rules and requirements defined by human experts.  
In practical implementation, however, rendering multi-view images of CAD models for validation purposes tends to be computationally expensive and slow, significantly impeding workflow efficiency.  
To address these limitations and facilitate effective learning from design failures, we propose the CJM by leveraging local compilers instead of relying on remote vision-language models (VLMs), which substantially expedite the validation process.

Specifically, as illustrated in Figure~\ref{fig:framework}, both the predicted sequence $\hat{y}$ and ground-truth CAD sequence $y$, represented as sketch-and-extrude (SE) parametric sequences, are first reconstructed into solid models. 
These solids are then converted to boundary representation (BRep) models, subsequently tessellated into meshes, and finally sampled uniformly to generate point clouds. 
The Chamfer Distance (CD) is computed between the predicted point cloud $\hat{P}$ (derived from $\hat{y}$) and the ground-truth point cloud $P$ (derived from $y$) as:  
\vspace{-4mm}
\begin{equation}
\text{CD}(\hat{P}, P) = \frac{1}{|\hat{P}|} \sum_{\hat{p} \in \hat{P}} \min_{p \in P} \|\hat{p} - p\|_2^2 + \frac{1}{|P|} \sum_{p \in P} \min_{\hat{p} \in \hat{P}} \|p - \hat{p}\|_2^2,
\end{equation}
where $|\hat{P}|$ and $|P|$ denote the number of points in $\hat{P}$ and $P$ respectively, and $\|\cdot\|_2$ represents the Euclidean distance. 
For samples that compile successfully and have CD below the threshold, we construct training tuples: with probability $\alpha$, we form $(x, \hat{y}, \text{true})$ to represent desired samples.
Samples failing to compile at any stage or with CD above a predefined threshold are marked as rejected that are with probability $1-\alpha$, we form $(x, y, \text{false})$ to represent undesired samples.  

By incorporating predicted sequences rejected in the first stage as ``rejection samples'', LLMs leverage such preference-derived data to rectify errors and reduce inaccuracies.  
Our approach constructs preference data by harnessing preference-based reward signals that are directly quantifiable through explicit rules from a CAD compiler, thereby alleviating the costs associated with the traditional rendering and ranking pipeline with VLM.  
Furthermore, leveraging local compilation tools and focusing on binary preferences enables efficient and scalable alignment.

\subsection{Compiler-as-a-Review Module (CRM) for Agentic CAD Generation}
Different from the CJM, the ground truth sequence is not accessible during the test phase. 
Therefore, CRM simply parses the sequence into the CAD model. 
The error information of the parsing would be utilized as the review of the sequence for self-refinement.

Specifically, after the model generates a sequence $\hat{y}$ from an input textual description, the output is immediately fed into a CAD compiler for validation. 
This validation process may identify various errors that cause compilation failures, including invalid sequence formatting (e.g., missing $\langle end\rangle$ terminators), geometric inconsistencies in sketch construction such as unclosable loops, incorrect extrusion parameters, and violations of boolean operation rules. 
Upon detecting an invalid sequence, the model performs re-sampling, combining the original input prompt $x$ with the error information derived from $\hat{y}$, leveraging its stochastic decoding mechanism to generate a new candidate sequence. 
The high efficiency of the local CAD compiler and compact tokenization of CAD commands ensure each validation step introduces negligible computational overhead, enabling our method to maintain real-time performance.

\subsection{Two Stage Fine-tuning for CAD Generation}
\textbf{Supervised Fine-Tuning for CAD Parametric Sequence Generation. }
We use a pre-trained LLM as the foundation of our method, aiming to leverage its accurate understanding of various language prompts and its knowledge of the 3D CAD design process~\cite{JSCAD}. 
We use the textual description introduced in Text2CAD \cite{text2cad}, and concatenate it with a fixed instruction to form the complete text input as $x$ to the LLM. 
We use the previously mentioned SE construction hierarchies as our textural CAD sequence $y$, which not only reduces the output difficulty and length for the LLM but also makes full use of the LLM's existing structured output capabilities. 

Our method fine-tunes the pre-trained LLM by minimizing the discrepancy between the generated parametric sequence $\hat{y}$ and the ground-truth parametric sequence $y$ using the cross-entropy loss function, denoted as
$\mathcal{L}_{SFT}$:
\begin{equation}
\mathcal{L}_{\text{sft}} = -\mathbb{E}_{(x,y) \sim D_{SFT}}\left[\frac{1}{T}\sum_{t = 1}^{T} \log p(\hat{y}_t = y_t|x)\right],
\end{equation}
where $T$ is the length of the sequence, and $p(\cdot) $is the predicted probability of the $t$-th token. 
By minimizing this loss function, we can make the generated parametric sequence closer to the ground-truth parametric sequence, thus improving the accuracy and reliability of the model in generating CAD parametric sequences.

\noindent\textbf{Preference Alignment with Binary Preference.}\\
Text-to-CAD generation can be inherently framed as an absolute preference task rather than a relative comparison task. 
These relative approaches often struggle to rank flawed outputs—particularly when such rankings cannot be directly derived from 3D models. 
Additionally, from a decision-theoretic perspective~\cite{tversky1992advances}, the expected utility hypothesis posits that rational agents evaluate actions based on their expected outcomes. 
In this context, the binary nature of rewards also enables models to learn from negative and positive samples with differentiated weights more effectively. 

To align large language models with preferences, the framework introduces an implied reward defined as:
\begin{equation}
    r_\theta(x, y) = \log \frac{\pi_\theta(y|x)}{\pi_{\text{ref}}(y|x)},
\end{equation}
with $\pi_\theta$ being the trainable model with parameters $\theta$, and $\pi_{\text{ref}}$ serving as the reference model.
A reference point $z_0$ is defined as the expected Kullback-Leibler divergence between the current model and reference model distribution:
\begin{equation}
    z_0 = \mathbb{E}_{x}[\text{KL}(\pi_\theta(\hat{y}|x) \| \pi_{\text{ref}}(\hat{y}|x))],
\end{equation}
serving as a divergence penalty to constrain distribution shift.

The value function is constructed as:
\begin{equation}
    v(x, y) = 
    \begin{cases} 
        \lambda_D \sigma(\beta(r_\theta(x, y) - z_0)) & \text{if } y \sim y_{\text{desirable}}|x, \\
        \lambda_U \sigma(\beta(z_0 - r_\theta(x, y))) & \text{if } y \sim y_{\text{undesirable}}|x,
    \end{cases}
\end{equation}
where $\lambda_D$ and $\lambda_U$ are weights assigned to desirable and undesirable samples respectively, $\sigma(\cdot)$ denotes the sigmoid function approximating the utility function, and $\beta$ is a scaling factor.
The KTO framework formulates the overall optimization objective as:
\begin{equation}
    L_{\text{KTO}} = \mathbb{E}_{x,y \sim D_{KTO}}[v(x, y)],
\end{equation}
where $v(x,y)$ represents the value function.
\section{Experiment}
\subsection{Setups}

\begin{table*}[t]
    \centering
    \caption{Quantitative test results on the DeepCAD dataset.
    The results include F1 scores for primitives and extrusions as well as mean and median Chamfer Distance (CD) and Invalidity Ratio (IR). 
    CD is multiplied by $10^3$.
    }
    \begin{tabular}{lcccccccc}
    \toprule
    \multirow{2}{*}{Prompt} & \multirow{2}{*}{Model} & \multicolumn{4}{c}{F1$\uparrow$} & \multicolumn{2}{c}{CD$\downarrow$} & \multirow{2}{*}{IR$\downarrow$}\\
    \cmidrule(lr){3-6} \cmidrule(lr){7-8}
    &  & Line & Arc & Circle & Extrusion & Median & Mean \\
    
    \midrule
    \multirow{5}{*}{L3} & DeepCAD~\cite{deepcad} & 0.77 & 0.20 & 0.65 & 0.89 & 32.82 & 97.93 & 10.00 \\
    & Text2CAD~\cite{text2cad} &0.81 &0.35 &0.74 &0.93 &0.37 &29.27 &2.38 \\
    & CADFusion~\cite{cadfusion} &0.79 &0.43 &0.69 &0.92 &- &30.23 &- \\
    & CAD-Coder~\cite{cad-coder} &- &- &- &- &0.17 &6.54 &1.45 \\
    \rowcolor{gray!20} & CAD-Judge (Ours) &\textbf{0.99} &\textbf{0.96} &\textbf{0.99} & \textbf{1.00} &\textbf{0.15} &\textbf{4.66} &\textbf{1.38}\\
    
    \midrule
    \multirow{3}{*}{L2} &Text2CAD~\cite{text2cad} & 0.73 & 0.07 & 0.66 & 0.93 & 74.1 & 150.01 & 1.84 \\
    &CADFusion~\cite{cadfusion} &0.67 &0.05 &0.48 &0.94 &- &146.15 &- \\
    \rowcolor{gray!20} &CAD-Judge (Ours) &\textbf{0.89}	&\textbf{0.49}	&\textbf{0.94}	&\textbf{1.00}	&\textbf{0.62}	&\textbf{52.55}	&\textbf{1.54} \\


    \bottomrule
\end{tabular}
\label{tab:deepcad}
\end{table*}
\begin{table*}[t]
    \centering
    \caption{Quantitative generalization results on the CADPrompt dataset and Fusion360 dataset.
    }
    \begin{tabular}{ccccccccc}
        \toprule
        \multirow{2}{*}{Dataset} &\multirow{2}{*}{Model} & \multicolumn{4}{c}{F1$\uparrow$} & \multicolumn{2}{c}{CD$\downarrow$} & \multirow{2}{*}{IR$\downarrow$}\\
        \cmidrule(lr){3-6} \cmidrule(lr){7-8}
        & & Line & Arc & Circle & Extrusion & Median & Mean \\
        
        \midrule
        \multirow{3}{*}{CADPrompt} &Text2CAD~\cite{text2cad} &0.67 &0.00 &0.31 &0.84 &127.70 &- &\textbf{1.57} \\
        &CADmium~\cite{cadmium} &0.69 &\textbf{0.22} &0.61 &0.89 &116.75 &- &6.28 \\
        \rowcolor{gray!20} &CAD-Judge (Ours) &\textbf{0.70}	&0.15	&\textbf{0.65}	&\textbf{0.90} 	&\textbf{42.69} &\textbf{152.40}	&2.55 \\

        \midrule
        \multirow{3}{*}{Fusion360} &Text2CAD~\cite{text2cad} &0.51 &0.14 &0.60 &0.87 &157.06 &239.51 &\textbf{3.57} \\
        &CADmium~\cite{cadmium} &\textbf{0.84} &\textbf{0.71} &0.90 &0.98 &76.05 &- &12.58 \\
        \rowcolor{gray!20} &CAD-Judge (Ours) &0.81	&0.61	&\textbf{0.91}	&\textbf{0.99} 	&\textbf{72.83} &\textbf{177.45}	&8.16 \\
        
        \bottomrule
    \end{tabular}
\label{tab:fusion360}
\end{table*}

\textbf{Datasets.}  
We utilize the widely adopted DeepCAD dataset~\cite{deepcad} as a source of CAD parametric sequences, with text prompts incorporated from Text2CAD~\cite{text2cad}.
The DeepCAD dataset is randomly partitioned into training, validation, and test sets at a ratio of 90\%--5\%--5\%, resulting in a total of ~137k samples aligned with the previous method.
The CADPrompt~\cite{cadprompt} comprises expert-written natural language instructions for 200 samples from DeepCAD dataset, serving as a crucial test for out-of-distribution generalization.
Additionally, we incorporate the more challenging Fusion360 Reconstruction dataset~\cite{fusion360}, paired with corresponding text descriptions derived from the CADmium dataset~\cite{cadmium}.
The Fusion360 dataset provides an official 80:20 split, consisting of 6,900 and 1,725 designs for training and testing, respectively.
All sketches and final CAD models are normalized to ensure consistency across samples.  
The parametric sequences are represented using 8-bit numerical precision, following prior work~\cite{deepcad, text2cad, cadfusion}, to facilitate stable and efficient training.

\textbf{Implementation Details.}  
Our experiments are conducted using PyTorch on a single NVIDIA RTX 4090 GPU. 
We apply Low-Rank Adaptation (LoRA)~\cite{lora} to enable efficient fine-tuning while maintaining performance. 
During SFT, the model is trained for 3 epochs with a batch size of 1, gradient accumulation over 8 steps, a learning rate of $ 1.0 \times 10^{-4} $, and a cosine scheduler with 10\% warm-up. 
In the second phase, outputs generated on the validation set are compared with ground-truth sequences to construct a binary preference dataset, with training performed at a reduced learning rate of $ 5.0 \times 10^{-6} $ for stability.

\textbf{Metrics.}  
Our evaluation focuses on the fidelity and alignment of generated CAD models with respect to input textual instructions. 
Following Text2CAD~\cite{text2cad}, we report three metrics: F1 score, Chamfer Distance (CD), and Invalidity Ratio (IR). 
The F1 score assesses correspondence between generated and ground-truth parametric sequences, computed separately for geometric primitives (lines, arcs, circles) and extrusion operations, then averaged. 
CD measures geometric similarity between point clouds of generated and reference 3D models, reflecting shape quality. 
Invalidity Ratio quantifies the fraction of syntactically invalid or non-executable outputs, indicating generation reliability.

\subsection{Main Results}
\textbf{Quantitative Evaluation.}
The results in Table~\ref{tab:deepcad} demonstrate the effectiveness of our proposed method, CAD-Judge, in generating CAD command sequences across different text prompts, compared to state-of-the-art approaches.

DeepCAD shows basic semantic understanding by identifying primitives and extrusions but lacks precise parameterization, resulting in poor geometric alignment~\cite{huang2023look}.
Text2CAD improves both semantic and geometric accuracy, reflecting better integration of language and shape understanding.
CADFusion excels in primitive classification through pre-trained language models but struggles with accurate parameter control, revealing a gap between recognition and execution.
CAD-Coder improves model reasoning by introducing a CoT planning process, demonstrating strong performance on the CD metric.
CAD-Judge achieves a better balance between semantic understanding and parametric precision. 
While it does not reach the lowest Invalidity Ratio (IR), this is largely due to inherent LLM output uncertainty. 
The CAD sequence structure extraction relies on closed-loop geometries sensitive to small errors, a problem worsened by strict CAD compiler constraints. 
Nonetheless, CAD-Judge significantly reduces IR compared to other LLM-based approaches such as CADFusion, while preserving high fidelity and validity.

In Table~\ref{tab:fusion360}, we evaluate the generalization capability of the model trained on DeepCAD across different datasets. 
It can be observed that our method significantly outperforms CADmium in both the CD metric and the IR metric. 
The slight underperformance in the ``arc'' may stem from the fact that when we utilize rule-based preference as the optimization signal, complex arcs become a ``high-cost option''. 
This incentivizes the model to adopt a shortcut strategy by preferentially using ``lines'' and ``circles'' to minimize CD value. 

\begin{figure}[t]
\centering
\includegraphics[width=\linewidth]{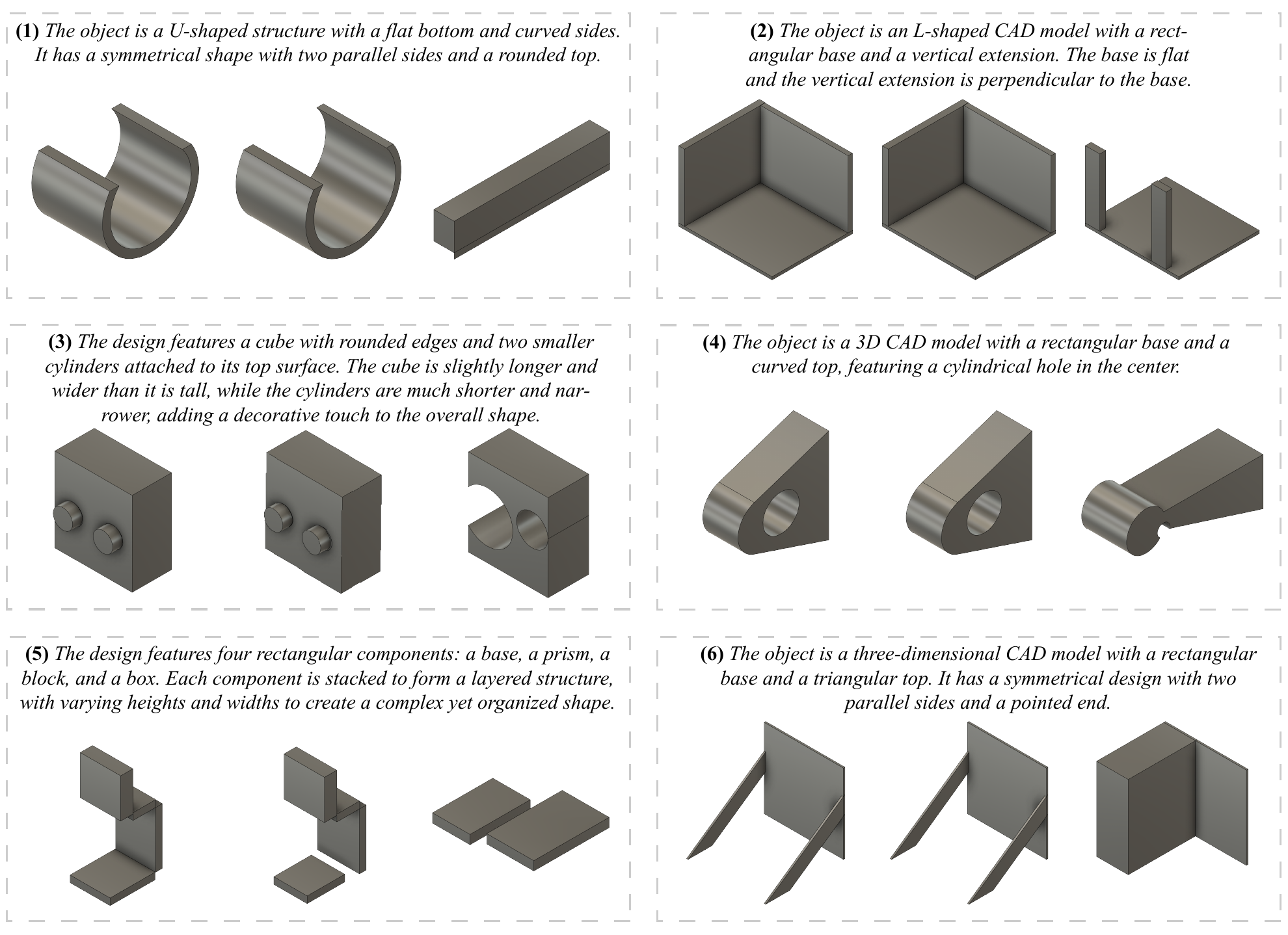}
\caption{Qualitative comparison of generated results. 
For each subsection, the input prompt is displayed at the top. 
From left to right, we show the ground truth, our CAD-Judge , and Text2CAD.}
\vspace{-4mm}
\label{fig:vis}
\end{figure}

\textbf{Qualitative Evaluation.}
Figure~\ref{fig:vis} compares the results among the ground-truth, our method, and Text2CAD on the test set. 
Text2CAD generate well-formed shapes without irregular edges or corners. 
However, it often produces oversimplified shapes and, for more complex prompts, tends to generate multiple cubes or panels instead of accurately capturing the intended structure. 
Our CAD-Judge provides the most precise response to input instructions and achieves the highest similarity to the ground truth. 
It successfully captures complex shapes, including rectangles, hexagons, and nested structures, such as a hexagonal hole within a cylinder. 
Additionally, it exhibits a strong understanding of language cues, accurately interpreting numerical and qualitative descriptors like ``long'' or ``T-shape''.

\begin{table*}[t]
    \centering
    \caption{Ablation studies on the proposed main components.
    Default settings are marked in gray.
    }
    \begin{tabular}{lccccccccc}
        \toprule
        \multirow{2}{*}{Exp} &Two Stage &Agentic & \multicolumn{4}{c}{F1$\uparrow$} & \multicolumn{2}{c}{CD$\downarrow$} & \multirow{2}{*}{IR$\downarrow$} \\
        \cmidrule(lr){4-7} \cmidrule(lr){8-9}
        &Fine-tuning &CAD Generation  & Line & Arc & Circle & Extrusion & Median & Mean \\
        \midrule
        1 &- &-          &0.89	&0.46	&0.93	&0.99	&0.91	&56.60	&3.07 \\
        2 &\checkmark &- &0.89	&0.47	&0.93	&\textbf{1.00}	&0.68	&52.96	&3.23\\
        3 &- &\checkmark &0.89	&0.47	&0.93	&\textbf{1.00}	&0.74	&55.12	&2.52\\
        \rowcolor{gray!20} 4 &\checkmark &\checkmark &\textbf{0.89} &\textbf{0.49}	&\textbf{0.94}	&\textbf{1.00}	&\textbf{0.62}	&\textbf{52.55}	&\textbf{1.54} \\
        \bottomrule
    \end{tabular}
\label{tab:ablation}
\end{table*}
\begin{table*}[t]
    \centering
    \caption{Ablation studies between preference data construction pipelines.
    Default settings are marked in gray.
    }
    \begin{tabular}{cccccccc}
        \toprule
        \multirow{2}{*}{Type}   & \multicolumn{4}{c}{F1$\uparrow$} & \multicolumn{2}{c}{CD$\downarrow$} & \multirow{2}{*}{IR$\downarrow$} \\
        \cmidrule(lr){2-5} \cmidrule(lr){6-7}
        &Line & Arc & Circle & Extrusion & Median & Mean \\
        \midrule
        VLM-based Paired Preference &0.76	&0.42	&0.76	&\textbf{1.00}	&18.52	&122.87	&6.02 \\
        Rule-based Paired Preference (Ours) &0.87	&0.48	&0.91	&\textbf{1.00}	&2.29	&69.58	&2.76  \\
        \rowcolor{gray!20} Rule-based Binary Preference (Ours) &\textbf{0.89}	&\textbf{0.49}	&\textbf{0.94}	&\textbf{1.00}	&\textbf{0.62}	&\textbf{52.55}	&\textbf{1.54} \\
        \bottomrule
    \end{tabular}
\label{tab:dpo}
\end{table*}

\subsection{Ablation Studies}
\textbf{Main Components.}
To dissect the contributions of our framework's core components, we perform an ablation study focusing on the proposed two-stage fine-tuning and agentic CAD generation, as shown in Table~\ref{tab:ablation}. By comparing Experiment 1 and Experiment 2, we observe that the introduction of two-stage fine-tuning leads to significant improvement in our model, primarily reflected in the CD metric. This outcome reveals that the two-stage preference dataset constructed through the CJM enhances the accuracy of parameters in the parameterized sequence, such as element positioning and extrusion size parameters. When comparing Experiment 1 and Experiment 3, it is evident that our proposed agentic CAD generation mechanism significantly reduces the model's invalid ratio. Furthermore, as this mechanism eliminates erroneous sequences, it ensures that the regenerated sequences are more parameter-consistent with real-world scenarios. Experiment 4 serves as our default setting. By combining the two judgment-based modules, it improves the accuracy of sequence parameters and reduces the invalid ratio in sequence generation.

\textbf{Preference Data Construction Comparison.}
We compared model performance across different preference data construction methods. 
The results in Table~\ref{tab:dpo} show that both rule-based construction methods outperform the VLM-based method. 
The VLM-based ranking uses GPT-4o to evaluate the consistency between rendered CAD images and text prompts for pairwise rankings. 
Rule-based ranking leverages CD metrics as reward signals to rank sequences, while our rule-based grading uses binary labels to indicate whether sequences meet the criteria. 
Rule-based ranking outperforms VLM-based methods in F1 (sequence element accuracy) and CD (model similarity) metrics, demonstrating the value of compiler-derived signals. 
Further gains from binary grading arise from reduced cognitive load and more consistent, noise-robust signals without data.

\begin{table}[t]
\centering
\captionof{table}{Ablation study on the numbers of iterations. Default settings are marked in gray.}
\begin{tabular}{lccc}
  \toprule
   \multirow{2}{*}{Iterations} & \multicolumn{2}{c}{CD$\downarrow$} & \multirow{2}{*}{IR$\downarrow$} \\
  \cmidrule(lr){2-3}
 & Median & Mean & \\
  \midrule
  0 &0.63	&53.11	&2.62 \\
  \rowcolor{gray!20} 1 &0.62 &52.55 &1.54 \\
  2 &\textbf{0.61} &\textbf{51.71} &1.45 \\
  3 &\textbf{0.61} &51.72 &\textbf{1.38} \\
  \bottomrule
\end{tabular}
\label{tab:crm}
\end{table}

\textbf{Iteration Limit of Agentic CAD Generation.}
We compare the iteration limit of agentic CAD generation. The results are summarized in Table~\ref{tab:crm}. Increasing the number of iterations steadily decreases the Invalidity Ratio (IR). However, the CD remains stable across different configurations. This observation suggests that while test-time rollback can reduce IR through compiler optimizations, increasing iterations does not lead to significant further improvements. We argue that excessive rollbacks are unnecessary. Therefore, in our implementation, we perform a single round of iteration.

\section{Conclusion}
We propose CAD-Judge, a novel alignment framework for text-to-CAD generation that rethinks the fine-tuning of LLMs in structured command synthesis. Our CAD-Judge leverages per-sample binary feedback through a Compiler-as-a-Judge Module, enabling efficient and interpretable training. We further introduce a Compiler-as-a-Review Module that monitors compilation outcomes and performs intelligent corrections upon detecting syntactic or geometric errors during the test phase. Our extensive experiments demonstrate that CAD-Judge outperforms existing approaches in both effectiveness and efficiency. Future work will explore reward-shaping strategies that incorporate semantic-level feedback beyond geometric fidelity, aiming to further enhance generalization and robustness.

\bibliography{aaai2026}

\end{document}